\documentclass[lettersize,journal]{IEEEtran}
\usepackage{amsmath,amsfonts}
\usepackage[ruled,vlined]{algorithm2e}
\usepackage{algpseudocode}
\usepackage{array}
\usepackage[caption=false,font=normalsize,labelfont=sf,textfont=sf]{subfig}
\usepackage{textcomp}
\usepackage{stfloats}
\usepackage{url}
\usepackage{verbatim}
\usepackage{graphicx}
\usepackage{cite}
\usepackage{amssymb}
\usepackage{amsmath}
\usepackage{booktabs}
\usepackage{hyperref}
\usepackage{subfig}
\usepackage{subfloat}
\usepackage{enumerate}
\usepackage{threeparttable}
\usepackage{color}
\usepackage{makecell}
\usepackage{multirow}
\usepackage{bm}
\begin{document}

\title{Pyramid Deep Fusion Network for Two-Hand Reconstruction from RGB-D Images}

\author{Jinwei~Ren,~and~Jianke~Zhu,~\IEEEmembership{Senior~Member,~IEEE} 
\thanks{Jinwei Ren and Jianke Zhu are both with the College of Computer Science and Technology, Zhejiang University, Zheda Rd 38th, Hangzhou, China. \protect\\
Email: \{zijinxuxu, jkzhu\}@zju.edu.cn;}
\thanks{Jianke Zhu is the Corresponding Author.}
}

\markboth{Journal of \LaTeX\ Class Files,~Vol.~14, No.~8, August~2021}%
{Shell \MakeLowercase{\textit{et al.}}: A Sample Article Using IEEEtran.cls for IEEE Journals}


\maketitle

\begin{abstract}
Accurately recovering the dense 3D mesh of both hands from monocular images poses considerable challenges due to occlusions and projection ambiguity. Most of the existing methods extract features from color images to estimate the root-aligned hand meshes, which neglect the crucial depth and scale information in the real world. Given the noisy sensor measurements with limited resolution, depth-based methods predict 3D keypoints rather than a dense mesh. These limitations motivate us to take advantage of these two complementary inputs to acquire dense hand meshes on a real-world scale. In this work, we propose an end-to-end framework for recovering dense meshes for both hands, which employ single-view RGB-D image pairs as input. The primary challenge lies in effectively utilizing two different input modalities to mitigate the blurring effects in RGB images and noises in depth images. Instead of directly treating depth maps as additional channels for RGB images, we encode the depth information into the unordered point cloud to preserve more geometric details. Specifically, our framework employs ResNet50 and PointNet++ to derive features from RGB and point cloud, respectively. Additionally, we introduce a novel pyramid deep fusion network (PDFNet) to aggregate features at different scales, which demonstrates superior efficacy compared to previous fusion strategies. Furthermore, we employ a GCN-based decoder to process the fused features and recover the corresponding 3D pose and dense mesh. Through comprehensive ablation experiments, we have not only demonstrated the effectiveness of our proposed fusion algorithm but also outperformed the state-of-the-art approaches on publicly available datasets. To reproduce the results, we will make our source code and models publicly available at {\url{https://github.com/zijinxuxu/PDFNet}}.

\end{abstract}


\section{Introduction}
Recovering the 3D pose and shape of human hands from a single viewpoint plays a pivotal role in a multitude of real-world applications, such as human-computer interaction~\cite{Song2012ContinuousBA}, mixed reality~\cite{Lin2010AugmentedRW}, action recognition~\cite{Fan2022UnderstandingAH}, and simulation. Over the past two decades, extensive research~\cite{zimmermann2017learning,romero2017embodied,Ge2018HandP3,boukhayma20193d,Moon2020I2L,Li2022InteractingAG,yu2023acr,Luan2023HighF3,Karunratanakul2022HARPPH,Xu2023H2ONetHN} has emerged in the field of hand reconstruction with various inputs including single color images, RGB-D images with depth maps, multi-view images, and video sequences. Due to the inherent complexity of finger joints, self-occlusions, and motion blur, an ongoing endeavor is to effectively address the challenges in 3D hand reconstruction.

\begin{figure}[t]
\centering
\includegraphics[width=0.98\columnwidth]{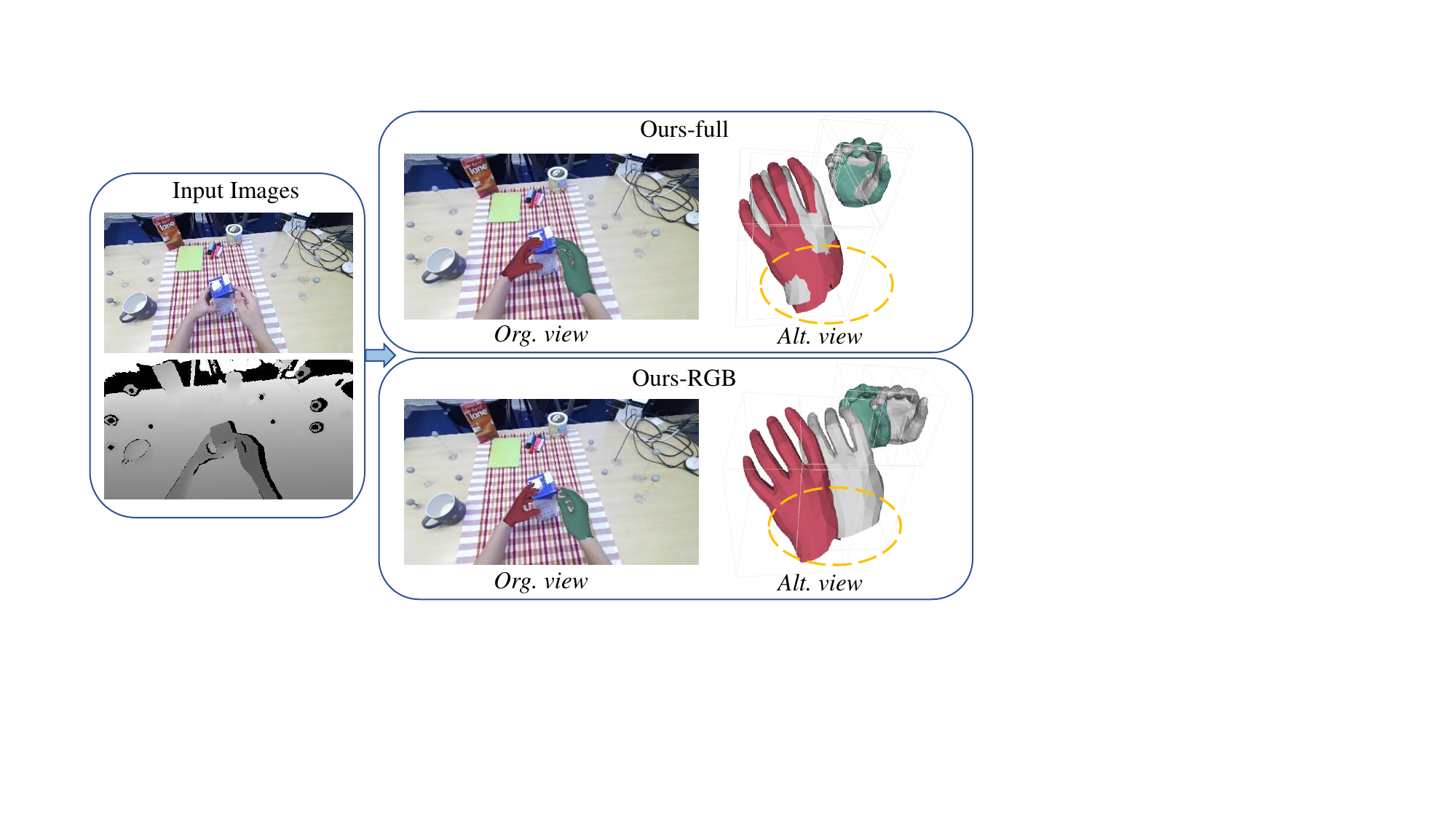} 
\caption{Comparison between Ours-full method and Ours-RGB method. Although the results of the two methods are very similar under the original projection perspective, there is a large misalignment of the latter in the depth direction under the new perspective.}
\label{example1}
\vspace{-0.2in}
\end{figure}

At present, the prevailing methods~\cite{Li2022InteractingAG,Zhang2021InteractingT3,Kim2021EndtoEndDA,Hampali2021KeypointTS} for hand reconstruction predominantly focus on directly estimating both hands from a single RGB image. However, these methods encounter difficulties in real-world scenarios characterized by cluttered backgrounds, lighting variations, and motion blur, which limit their performance to environments similar to the training data. Generally, a conventional framework~\cite{zimmermann2017learning} separates detection and reconstruction, which requires extracting the hand region from the image by an off-the-shelf detector before feeding it to the reconstruction model. Consequently, these models only predict root-aligned 3D hand meshes. Instead, depth map-based methods~\cite{cai2018weakly,Cai20213DHP} often incorporate range maps as auxiliary supervisory information to compensate for inherent noise and limited resolution. Additionally, certain approaches~\cite{Ge2018HandP3,Li2018PointToPoseVB} employ depth maps to predict sparse 3D keypoints. The absolute scale information in depth maps is not affected by background changes as well as the rich foreground features in RGB maps, which is crucial to hand reconstruction. Fig.~\ref{example1} presents a visual comparison between utilizing solely RGB input and augmenting it with depth map input. 

The previous fusion methods~\cite{Xu2017PointFusionDS,Chen2020SpatialIG,Wu2022RobustRF} have primarily relied on RGB-D cameras, which leverage both rich image information and depth measurements to accomplish tasks such as object detection and semantic segmentation. Despite extensive research efforts over an extended period, an effective fusion scheme for hand reconstruction remains elusive. This challenge can be attributed to the highly nonlinear nature of gestures~\cite{Cheng2016SurveyO3} and the inherent variations between hands, making it arduous to achieve satisfactory results through a straightforward combination of color images and depth maps. In certain scenarios, utilizing depth maps alone can actually yield superior outcomes~\cite{kazakos2018fusion}. Hence, it becomes imperative to ascertain an effective fusion strategy specifically tailored for hand reconstruction tasks.

The simplest and most rudimentary fusion method entails directly incorporating the depth map as an additional channel alongside the RGB image~\cite{Mueller2017RealTimeHT}. This approach merely requires modifying the input channel of the model from three to four channels. However, the performance enhancement achieved by this simple fusion method remains quite limited. An alternative fusion approach that has gained popularity is operating at the feature level~\cite{kazakos2018fusion,Lin2021MultiLevelFN,Sun2021CrossFuNetRA}. It is important to note that directly concatenating the two features obtained from shallow CNNs does not yield any performance improvements~\cite{kazakos2018fusion}. Accordingly, researchers have made endeavors to extract multi-scale depth features~\cite{Lin2021MultiLevelFN} or perform cross-fusion at intermediate feature layers~\cite{Sun2021CrossFuNetRA}.

The aforementioned fusion methods are primarily employed for the cropped single-hand images, which are limited to predicting sparse 3D keypoints rather than dense meshes. Furthermore, these methods process depth maps into 2D images, disregarding their inherent 3D characteristics. Inspired by previous work in 3D object detection~\cite{Xu2017PointFusionDS} and 6-DoF estimation~\cite{Wang2019DenseFusion6O,Wang2022EFN6DAE}, we adopt a different approach by converting the depth map into an unordered point cloud, and then extract point features to fuse them with RGB features derived by CNNs. Experimental results indicate that this method yields the improved performance due to the more effective feature. Additionally, we argue that it is insufficient for learning local features by relying solely on fixed-sized point feature regression. Motivated by the architecture of PointNet++~\cite{Qi2017PointNetDH}, we introduce a pyramid feature fusion module that enable the integration of point cloud features and RGB features at their corresponding positions across multiple scales. Moreover, existing frameworks based on sparse 3D keypoints or root-aligned mesh estimation may fall short when attempting to achieve two-hand reconstruction in real-world interactive scenarios.

In order to address the aforementioned challenges, we present an end-to-end framework that incorporates RGB and depth information to accurately reconstruct a 3D mesh of both hands from RGB-D inputs. Unlike HandPointnet~\cite{Ge2018HandP3}, our approach eliminates the need for normalizing the point cloud using oriented bounding boxes, thereby avoiding misalignment between the point cloud and color image while simplifying the process.
To tackle the difficulty in learning local features, we suggest a pyramid structure feature fusion module called PDFNet, which facilitates the fusion of two features at different scales in order to enable the effective integration of information. Furthermore, we introduce an adaptive weight allocation module to achieve more robust and accurate fusion, which allocates weights to different features to mitigate interference from local unreliable regions.

To attain a more comprehensive representation of hand structures, as opposed to merely sparse 3D keypoints, we opt to employ a graph convolutional network (GCN) as our decoder as in~\cite{Li2022InteractingAG}. Instead of directly using the image-wide features, we introduce a center map for precise hand localization. Additionally, we conduct experiments using the parameterized model MANO~\cite{romero2017embodied} and multiple fully connected layers as alternative decoders across various two-hand datasets. The results demonstrate the convincing performance enhancement achieved through our fusion algorithm.

From above all, the main contributions of our work can be summarized as follows.
\begin{enumerate}[(1)]
\item  We propose an efficacious end-to-end single-stage framework that reconstructs 3D hand meshes from a solitary RGB-D input. To the best of our knowledge, this is the first RGB-D fusion framework for two-hand reconstruction. 
\item We devise a novel fusion module named PDFNet that effectively harnesses both color information and depth maps. Empirical studies validate the substantial enhancement that this module imparts upon the baseline model.
\item Both quantitative and qualitative evaluations clearly demonstrate that our proposed approach achieves state-of-the-art performance on publicly available two-hand datasets~\cite{Kwon2021H2OTH, Hampali2021KeypointTS}.
\end{enumerate}

\section{Related Work}
Rapid progress has been made on hand pose estimation~\cite{Wang2019MaskPoseCC,zimmermann2017learning,Guo2021GraphBasedCW,Ge2018HandP3,Ge2018PointtoPointRP,Li2021LatentD3} and 3D hand mesh~\cite{romero2017embodied,Moon_2020_ECCV_DeepHandMesh,Cai20213DHP,Lee2023Im2HandsLA} reconstruction over recent years, giving rise to various categories such as single-handed~\cite{boukhayma20193d,Guo20223DHP} and multi-handed reconstruction~\cite{Ren2023EndtoEndWS}, fully supervised ~\cite{Moon2020I2L,Lin2021MeshG} and weakly supervised methods~\cite{Kulon2020WeaklySupervisedMH,Spurr2021SelfSupervised3H,Tu2022Consistent3H}, etc. In this paper, our primary research focus lies in exploring different types of inputs. Consequently, previous studies can be classified into three distinct groups, namely color image, depth map, and RGB-D image.

\subsection{Hand Reconstruction from Color Image}
Due to the lack of depth information, it is very challenging to recover 3D hand pose from a single color image. Zimmermann \textit{et al}.~\cite{zimmermann2017learning} trained a deep neural network to learn the 3D articulation prior of hands on a synthetic dataset. Guo \textit{et al.}~\cite{Guo20223DHP} proposed a feature interaction module to enhance the joint and skeleton feature. In addition to predicting 3D pose, Boukhayma \textit{et al.}~\cite{boukhayma20193d} further predicted the shape of the hand and optimized the 3D parameterized model MANO~\cite{romero2017embodied} through a re-projection module. To improve performance, the subsequent model-based methods introduced iterative optimization~\cite{zhang2019end}, neural rendering~\cite{baek2019pushing}, spatial mesh convolution~\cite{Kulon2020WeaklySupervisedMH}, adaptive 2D-1D registration~\cite{Chen2021CameraSpaceHM}, etc. 
In addition, novel image-to-pixel prediction networks~\cite{Moon2020I2L}, graph-convolution-reinforced transformer~\cite{Lin2021MeshG}, and contrastive learning~\cite{Spurr2021SelfSupervised3H} have also been applied in this field. In order to address the scarcity of 3D annotations for real hands, Zimmermann \textit{et al.}~\cite{zimmermann2019freihand} proposed the first single-hand dataset containing 3D pose and shape labels. Hampali \textit{et al.}~\cite{hampali2020honnotate} proposed a dataset with similar annotations, which focuses on hand-object interaction scenes. Considering the situation of multiple hands in a picture, multi-stage methods~\cite{Panteleris2018UsingAS}~\cite{rong2021frankmocap} that separate hand detection and pose estimation, as well as single-stage methods~\cite{Ren2023EndtoEndWS}~\cite{yu2023acr} that jointly detect and reconstruct, have been proposed. Moon \textit{et al.}~\cite{moon2020interhand2} proposed a large-scale real-captured interacting hand dataset using a multi-view system. Based on this dataset, several subsequent works~\cite{Li2022InteractingAG,Zhang2021InteractingT3,Kim2021EndtoEndDA,Hampali2021KeypointTS,Wang2023MeMaHandEM} have conducted more in-depth research on left and right hand interaction and designed exquisite network structures to better extract features.

\subsection{Hand Reconstruction from Depth Map}
Compared to using only RGB images, it is more intuitive to recover the hand pose and shape from depth maps, as partial geometric information can be directly obtained. According to the different processing methods for input data, these methods can be roughly divided into two categories, including image-based methods and point cloud-based approaches. The former mostly directly employs CNNs to process depth maps like RGB images through feedback loop~\cite{Oberweger2015Training}, dense per pixel compression~\cite{Wan2017Dense3R}, forward kinematics~\cite{zhou2016model}, adaptive weighting regression~\cite{Huang2020AWRAW}, and auxiliary latent variable~\cite{Xu2021ImproveRN}, etc. The latter processes the depth map into a point cloud and directly extracts point features from it to regress the hand pose. Ge \textit{et al.}~\cite{Ge20173DCN} proposed 3D CNN for point feature extraction and regress full hand pose in volumetric representation. In order to effectively utilize information in depth images and reduce network parameters, Ge \textit{et al.}~\cite{Ge2018HandP3} adopted the network structure of PointNet~\cite{Qi2016PointNetDL,Qi2017PointNetDH} to extract point cloud features. The point cloud regularization module is also introduced to improve the robustness of the method. Subsequent work adopted similar frameworks while introducing the intermediate supervisory information such as heatmap and unit vector field~\cite{Ge2018PointtoPointRP}, semantic segmentation~\cite{Chen2018SHPRNetDS} to enhance the performance of the model. With the deepening of research, permutation equivariant layer (PEL)~\cite{Li2018PointToPoseVB}, self-organizing map (SOM)~\cite{Chen2019SOHandNetSN} and Transformer~\cite{Huang2020HandTransformerNS} have also been introduced into hand pose estimation. As for interacting hand, Taylor \textit{et al.}~\cite{taylor2017articulated} trained a segmentation network to construct a 3D point cloud from depth maps and designed a signed distance field to minimize model fitting errors. Muller \textit{et al.}~\cite{mueller2019real} estimated a vertex-to-pixel correspondence map first and proposed an energy minimization framework, which can optimize the pose and shape parameters by fitting the point cloud. However, optimization-based model fitting methods rely more on precise 3D point cloud inputs. On the other hand, the learning-based methods may obtain more prior information from other depth maps to mitigate the impact of sensor noise, which has not yet been explored. 

\begin{figure*}[ht]
\centering
\includegraphics[width=0.96\textwidth]{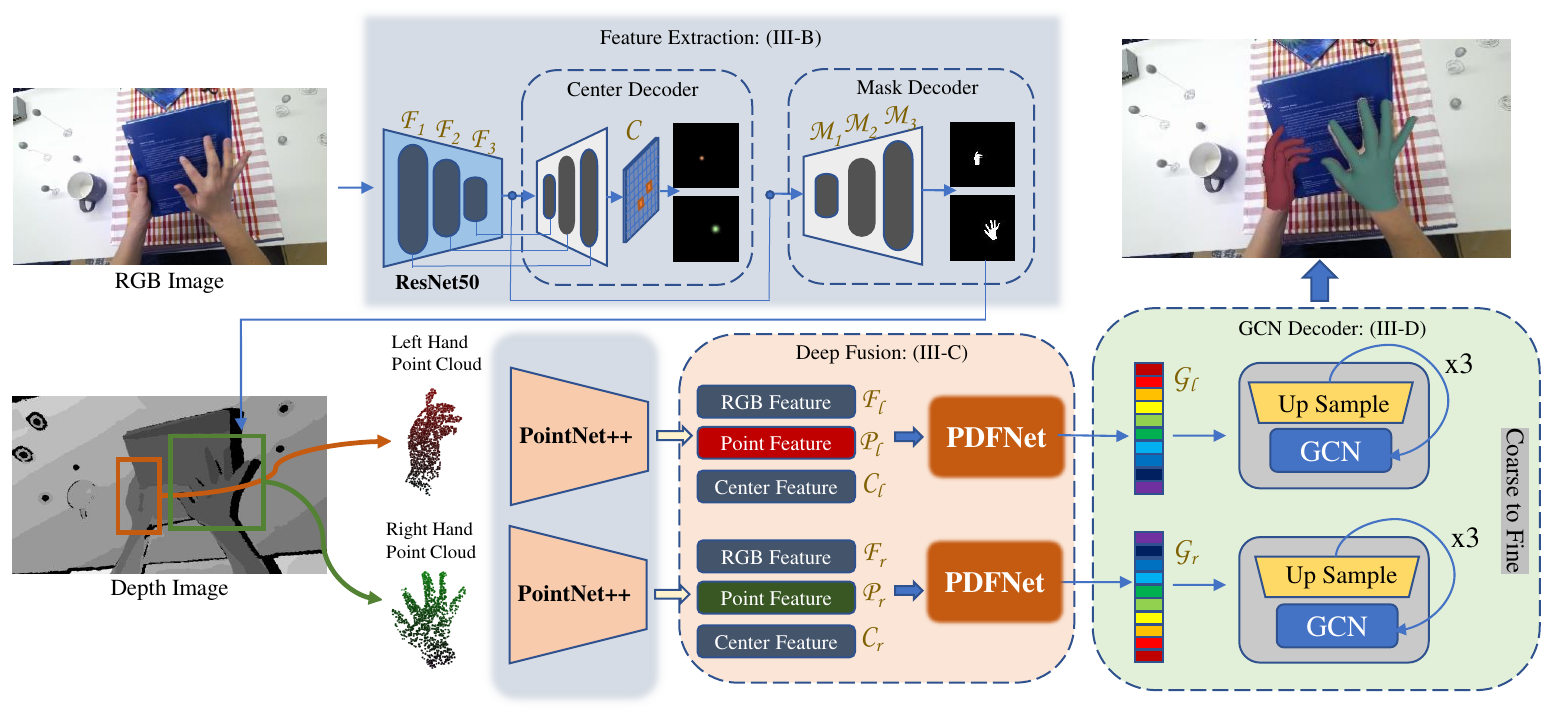} 
\caption{Overview of the proposed framework. Given an RGB-D image, we adopt ResNet50~\cite{He2016DeepRL} and PointNet++~\cite{Qi2017PointNetDH} as the backbone to extract features (Section~\ref{sec:encoder}) and decode RGB features into center maps and masks using two simple decoders. The deep fusion module (Section~\ref{sec:pdfnet}) is responsible for the deep fusion of RGB features and point features. The GCN-based decoder (Section~\ref{sec:decoder}) takes the fused global feature and outputs dense hand mesh of both hands in a coarse to fine way. The whole pipeline is trained in an end-to-end manner.}
\label{framework}
\vspace{-0.2in}
\end{figure*}

\subsection{Hand Reconstruction from RGB-D Image}
RGB-D fusion has been extensively studied in fields such as 3D object detection~\cite{Xu2017PointFusionDS}, object pose estimation~\cite{Wang2019DenseFusion6O}, and semantic segmentation~\cite{Wu2022RobustRF}~\cite{Chen2020SpatialIG}. However, there has been few in-depth research on hand reconstruction tasks. ~\cite{Oikonomidis2012TrackingTA}~\cite{Sridhar2016RealTimeJT}~\cite{Tzionas2015CapturingHI} requested RGB-D sensor as input while the RGB image was only used to segment the hand part in the depth map. Cai \textit{et al.}~\cite{Cai20213DHP} used depth maps as regularization terms during training to reduce dependence on 3D annotations, and only used RGB images during testing. Yuan \textit{et al.}~\cite{Yuan20193DHP} pre-trained a depth-based network and froze the parameters of the network during joint RGB-D training. The gap between the RGB-based method and the depth-based approach is narrowed by minimizing the intermediate features of the two branches. Kazakos \textit{et al.}~\cite{kazakos2018fusion} designed a double-stream architecture for RGB-D fusion, and tried input-level fusion, feature-level fusion, and score-level fusion. Unfortunately, their experiments indicated that adding RGB information did not help with performance gains. Mueller \textit{et al.}~\cite{Mueller2017RealTimeHT} directly used the 4-channel RGB-D input and trained two CNNs to locate and regress the 3D position of the hand. They chose to project RGB pixels onto a depth map to obtain a colored depth map and then predicted the absolute coordinates of the hand center and the 3D offset of each joint separately. Lin \textit{et al.}~\cite{Lin2021MultiLevelFN} scaled the depth map to multiple sizes to aggregated features at different resolutions, and then adopted feature attention structures~\cite{Hu2017SqueezeandExcitationN} to fuse RGB features. Sun \textit{et al.}~\cite{Sun2021CrossFuNetRA} adopted a similar dual stream structure, where the depth map branch used a shallower network to avoid overfitting. The features of the two branches were first cross fused in the middle part, and then concatenated together. This resulted in better results compared to direct concatenation. 

Each of the aforementioned approaches exhibits certain limitations. Primarily, they solely focus on regressing hand pose without undertaking shape reconstruction. While point cloud structures offer a more accurate depiction of geometric information compared to depth maps, there is a dearth of research in this area. Additionally, simply stitching the global features may pose challenges in effectively capturing local structures. A potential solution lies in multi-scale feature fusion. By taking into account these aforementioned limitations, we introduce a novel framework for dense hand mesh reconstruction, built upon our pyramid fusion module (PDFNet).

\section{Methodology}
The goal of this paper is to restore a dense 3D mesh of both hands within real-world scenes through a single RGB-D image. Our framework takes both RGB images and point cloud generated from depth maps as inputs, and extracts features using classic ResNet50~\cite{He2016DeepRL} and PointNet++~\cite{Qi2017PointNetDH}, respectively.
Subsequently, the extracted features are fed into PDFNet for deep fusion to improve the performance of our model. The fused features are then fed into the GCN-based decoder to output the dense 3D mesh of both hands. By ingeniously fusing the modalities of RGB and depth, we are able to accurately reconstruct a 3D hand mesh with real depth and scale in camera space. For interactive scenarios in AR/VR applications, it is imperative to restore the absolute position within the camera coordinate system, surpassing the limitations of previous root-aligned outputs. Apart from the root position, the depth map also conveys the relative geometric relationship among hand joints, which yields significant contributions to the accuracy of reconstruction in local coordinate systems.
\subsection{Overview}
The overall structure of our approach is  a classical encoder-decoder architecture, as depicted in Fig.~\ref{framework}. Our method can be divided into three integral components, including feature extraction, feature fusion, and feature decoding. Within the feature extraction module (Section~\ref{sec:encoder}), we extract 2D image features utilizing ResNet50, while simultaneously extracting 3D point cloud features using PointNet++. In the feature fusion phase (Section~\ref{sec:pdfnet}), the corresponding RGB features and point cloud features are fused at the pixel level through point cloud indexing. Finally, in the feature decoding phase (Section~\ref{sec:decoder}), we employ multi-layer Graph Convolutional Networks (GCN) and upsampling operations to decode the input global features into a finely detailed 3D dense mesh representation with two distinct hands. In the following sections, we will provide a detailed description of each module in the framework.

\subsection{Dual-stream Encoder}
\label{sec:encoder}
In the feature extraction module, we need to fully extract the features from the RGB-D image containing both hands from the first perspective to restore accurate pose and shape. 

\noindent \textbf{RGB Feature Extraction.} Firstly, given an unprocessed monocular RGB image $\mathcal{I}_{c} \in \mathbb{R}^{{H}\times{W}\times{3}}$, we use the classic ResNet50 to extract 2D pyramid features as follows: $F = \{
{\mathcal{F}_{1}} \in \mathbb{R}^{{H} \times{{W} \times{3}}}, {\mathcal{F}_{2}} \in \mathbb{R}^{{\frac{H}{2}} \times{\frac{W}{2}} \times{64}}$,${\mathcal{F}_{3}} \in \mathbb{R}^{{\frac{H}{4}} \times{\frac{W}{4}} \times{256}} \}$. Then we adopt two simple decoder networks to regress the center $P_{ct} = \{ P_{l} \in \mathbb{R}^2, P_{r} \in \mathbb{R}^2 \}$ and mask $M = \{ M_{l} \in \mathbb{R}^{{H} \times{{W}}}, M_{r} \in \mathbb{R}^{{H} \times{{W}}} \}$ of the left and right hands. The predicted center position of each hand will be used to initialize the 3D position of the hand mesh, and the predicted mask will be used to segment the hand area in the depth map.

\noindent \textbf{Point Cloud Preprocessing.} Given an unprocessed depth map $\mathcal{I}_{d} \in \mathbb{R}^{{H}\times{W}\times{1}}$ and predicted mask $M$ for both hands, we first convert the 2D image into a 3D point set using the camera's parameters. By calculating the mean depth of the point set, we filtered out outliers that exceed the threshold range [-0.08,+0.08] mm to reduce noise interference. Then, we randomly selected 1024 points from the remaining point set as the initial point cloud. Based on the generated initial point cloud, we can directly extract point cloud features using a specially designed network.

\noindent \textbf{Review of PointNet++.}
Compared to directly extracting features from 2D depth maps using CNNs, PointNet~\cite{Qi2016PointNetDL} pioneered the extraction of high-dimensional features directly from point cloud through a per-point multi-layer perceptron (MLP) network. However, there is a lack of mining for local structural features due to the fixed number of points in PointNet. Therefore, PointNet++~\cite{Qi2017PointNetDH} proposed a hierarchical feature extraction architecture to address this issue. Specifically, it includes multiple point set abstraction levels by selecting a fixed number of points in each layer as the center of the local area. The K neighbors around each center point will be aggregated and high-dimensional features will be extracted through the classic PointNet network. The center point and high-dimensional features will be fed into the next layer and the aggregation operation will be repeated. Finally, global features are extracted from all points in the last layer through the PointNet network. It is worthy of noting that previous work often used PointNet for point cloud classification, and it is still an unexplored field to predict dense 3D meshes from sparse point cloud. 

\noindent \textbf{Depth Feature Extraction.}
Given a set of point cloud data with both hands $X_{h} = \{ X_{l} \in \mathbb{R}^{{N}\times{C}}, X_{r} \in \mathbb{R}^{{N}\times{C}} \}$, we refer to the structure of PointNet++ to extract pyramid point cloud features as follows: $ P = \{{\mathcal{P}_{1}} \in \mathbb{R}^{{2} \times{{N} \times{C}}}, {\mathcal{P}_{2}} \in \mathbb{R}^{{2} \times{{N_{1}}} \times{C_{1}}}$,${\mathcal{P}_{3}} \in \mathbb{R}^{{2} \times{{N_{2}}} \times{C_{2}}} \} $. In our implementation, $N$=1024, $C$=3, $N_{1}$=512, $C_{1}$=131, $N_{2}$=128, $C_{2}$=259. At each level of point set abstraction, our approach involves employing ball queries to locate neighboring points within a predefined radius range. These identified points are subsequently fed into Multi-Layer Perceptrons (MLPs), enabling the extraction of high-dimensional features that correspond to the number of central points. These features are then concatenated with the features of the central point, yielding the point cloud features specific to that particular layer. Note that the pyramid point cloud features obtained at this stage exhibit a structure similar to the pyramid RGB features obtained earlier. In other words, as the scale decreases, the number of channels deepens, representing a continuous process of feature abstraction. By integrating pyramid features at different scales, we delve further into the feature characteristics of distinct modalities, thereby mutually reinforcing and complementing each other.

\begin{figure*}[ht]
\centering
\includegraphics[width=0.96\textwidth]{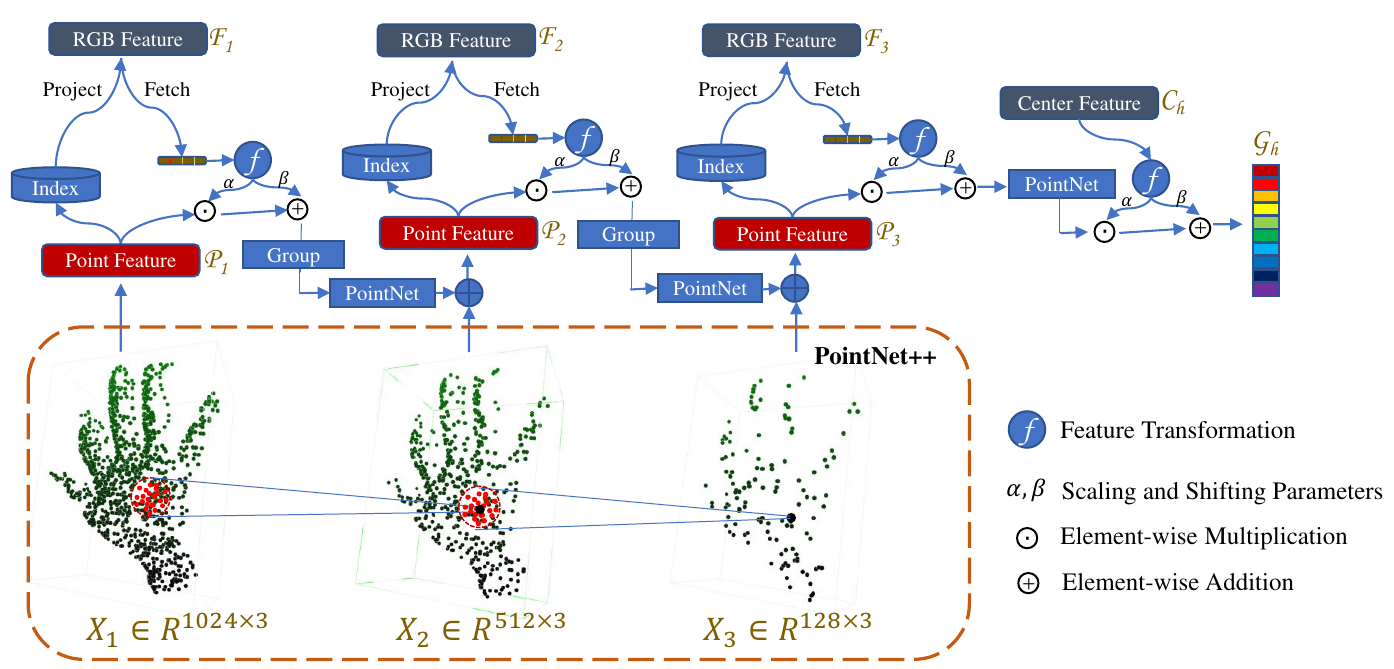} 
\caption{Details of our proposed Pyramid Deep Fusion Network (PDFNet).}
\label{PDFNet}
\vspace{-0.2in}
\end{figure*}

\subsection{Pyramid Deep Fusion Network}
\label{sec:pdfnet}
At this stage, we have successfully obtained pyramid features for both modalities. Now, the crucial step is to fuse these features effectively. While the most simplistic approach involves a single layer of MLP to generate global features from the two modalities, followed by their concatenation, such a method neglects the local discrepancies present between the modalities. Factors such as motion blur, occlusion, and noise are significant local characteristics that might impair the ability of global features to complement each other. To address this issue, we have adopted a pixel-level feature fusion technique, aligning the corresponding RGB features with the point cloud features through 3D-2D projection of the point cloud. Notably, unlike the approach employed in DenseFusion~\cite{Wang2019DenseFusion6O}, we perform pixel-by-pixel fusion on multi-scale pyramid features. In contrast to simply concatenating two distinct features, we incorporate a feature space transformation module. This module dynamically allocates weights to avoid the influence of local biases on the overall performance.

Specifically, we have designed a three-layer pyramid feature fusion structure, as shown in Fig.~\ref{PDFNet}. With the help of PointNet++~\cite{Qi2017PointNetDH} network, we downsample the initial point cloud ${\mathcal{X}_{1}} \in \mathbb{R}^{{1024} \times{{3} }}$ to more sparse point cloud ${\mathcal{X}_{2}} \in \mathbb{R}^{{512} \times{{3} }}$ and ${\mathcal{X}_{3}} \in \mathbb{R}^{{128} \times{{3} }}$ through central point aggregation. Each set of point cloud finds K neighboring points as a local point set through the ball query of the center point. Through the PointNet network, higher-dimensional features are extracted from each local point set and subsequently consolidated into a single point representation via max-pooling. The resulting aggregated high-dimensional features are concatenated with the center point features from the original point cloud to obtain the point features $P$ of that layer. For easier comprehension, the corresponding pseudocode can be seen in algorithm~\ref{alg1}.

To acquire the RGB features corresponding to specific positions, we retain the index vector of the point cloud with respect to the depth map. Through this index, we project each point cloud onto a 2D feature map and gather the corresponding features, as illustrated in the projection-fetch process depicted in Fig.~\ref{PDFNet}. The collected RGB features and point cloud features possess similar dimensions to facilitate seamless feature stitching, which is deemed as the conventional and effective approach for feature fusion. However, disparities exist in the data distribution and magnitude order between the two feature vectors, prompting the need for adaptive allocation of feature weights in order to attain enhanced outcomes. Motivated by the Spatial Feature Transform technique introduced by~\cite{Wang2018RecoveringRT}, we have tailored a shallow MLP network to learn scale and shift parameters individually for the aforementioned two features. The RGB features serve as the conditioning factor to acquire the scale and shift parameters. This enables a feature affine transformation that maps the point features into a novel feature space, as illustrated below
\begin{equation}
\mathcal{\hat{P}} = \mathcal{P} \odot \bm{\alpha} + \bm{\beta},  (\bm{\alpha}, \bm{\beta}) = \psi(\mathcal{F}).
\end{equation}
$\bm{\alpha}$ and $ \bm{\beta}$ are learned affine transformation parameters scale and shift, whose dimension is the same as $\mathcal{P}$. $\odot$ refers to element-wise multiplication, while $\psi$ refers to our feature transformation network. The transformed feature $\mathcal{\hat{P}}$ will be aggregated into a more sparse high-dimensional feature point cloud through the point set abstraction layer of PointNet++. The point cloud features of the last layer are fused to generate a single global feature $\mathcal{G} \in \mathbb{R}^{{2} \times{1024} \times{1}}$ through PointNet network. After obtaining the fused features, we aim to merge them with the center features derived from CNNs. The center features represent the global characteristics of the entire hand, while the fused features consist of sparse local features. This design combines global and local elements, which maximizes the representation of input images and leads to improved results. Our subsequent ablation experiment further corroborates this discovery. Once we have acquired the final fused features, they are fed into our GCN-based decoder to output dense 3D meshes of both hands.

\begin{algorithm}[t]
\caption{Algorithm for PDFNet Procedure }\label{alg1}
\KwIn{pyramid RGB feature map $F$; center feature map $C$; initial point cloud $X_{h}$; camera intrinsic matrix $K$; $num\_layers$; $BallRadius$; $NumPoints$.}
\KwOut{global fused feature $\hat{G}$.}
$P_{1}$, $X_{1}$ $\leftarrow$ $X_{h}$  \Comment{Initialize point feature and point set}

\For{$i$ in $[1, num\_layers]$ }{ 
$(u,v)$ $\leftarrow$ $K^{-1}X_{i}$ \Comment{Find image coordinates}\\
$\hat{F_{i}}$ $\leftarrow$ $Fetch(F_{i}|u,v)$  \Comment{Fetch corresponding RGB features}\\
$(\alpha, \beta)$ $\leftarrow$ $\psi_{i}(\hat{F_{i}}, P_{i})$ \Comment{Calculate affine transformation parameters}\\
$\hat{P_{i}}$  $\leftarrow$ $(P_{i} \odot (\alpha + 1) + \beta)$ \Comment{Feature transformation}\\

\eIf{$i < num\_layers$}{
$S_{i}$ $\leftarrow$ find($\hat{P_{i}},NumPoints\_{i},BallRadius\_{i}$) \\ \Comment{Find local structure}\\
$P_{i+1}$ $\leftarrow$ $cat(P_{i},PointNet(group(S_{i})))$ \\  \Comment{Point set abstraction}\\
}
{
$G$ $\leftarrow$ $PointNet(\hat{P_{i}})$  \\ 
$(\alpha, \beta)$ $\leftarrow$ $\psi_{i+1}(C, G)$ \\ 
$\hat{G}$ $\leftarrow$ $(G \odot (\alpha + 1) + \beta)$ \Comment{Affine transformation} \\
}

}
\Return{$\hat{G}$}
\end{algorithm}
\subsection{GCN-based Decoder}
\label{sec:decoder}
To fully leverage the extracted feature information, our decoder has been primarily constructed on the foundation of a state-of-the-art (SOTA) method~\cite{Li2022InteractingAG}. Unlike primarily addressing interactive hands positioned at the center, our method accommodates hands appearing in any position within the field of view. Thus, we draw inspiration from the design of CenterNet~\cite{zhou2019objects} and utilize the center point as a representation of the hand. When extracting the corresponding image features, we collect global features from the central point position within the feature map, instead of directly flattening the entire map. Subsequent comparative experiments have substantiated the advantages of our approach, as it effectively focalizes the features on the hand regions rather than the background areas, ultimately yielding superior results.

We employ the Chebyshev Spectral Graph Conventional Network~\cite{Defferrard2016ConvolutionalNN} to construct our 3D hand mesh, following the classic Coarse-to-fine structure. As in~\cite{Li2022InteractingAG}, we construct a three-layer submesh with designated vertex quantities, $N_{1}$=63, $N_{2}$=126, $N_{3}$=252. The final mesh is consistent with the topology of MANO~\cite{romero2017embodied}, containing 778 vertices. Leveraging multiple upsampling layers, we successfully refine the hand mesh from the initial coarser submesh to the ultimate full MANO mesh.

Similar to PointNet, GCN learns the geometric structure of 3D meshes by directly optimizing the features on each vertex. Given the fused global feature $\mathcal{G}$, we map it into a more compact feature vector through a fully connected layer and concatenate it with the position encoding of vertices to obtain our initial graph features $\mathcal{G_{V}} \in \mathbb{R}^{{N}\times {C}}, (N=63,126,252),(C=512,256,128)$. Similar to ~\cite{Li2022InteractingAG}, our graph convolution operation on each graph feature is defined as follows:
\begin{equation}
{G_{out}} = \sum_{k=0}^{K-1} C_{k}(\hat{L}) G_{in} W_{k}.
\end{equation}
where $C_{k}$ is Chebyshev polynomials of degree k and $\hat{L} \in \mathbb{R}^{{N}\times {N}}$ is the scaled Laplacian matrix. $W_k \in \mathbb{R}^{{C_{in}}\times {C_{out}}}$ is a learnable weight matrix. $G_{in} \in \mathbb{R}^{{N}\times {C_{in}}}$ and $G_{out} \in \mathbb{R}^{{N}\times {C_{out}}}$ are input and output features in graph convolution operations, respectively. Through multiple regression heads composed of fully connected layers, we map the graph features of the last layer to the corresponding optimization objectives, such as root node coordinates, root-aligned MANO mesh, GCN mesh, etc.

\subsection{Loss Functions} 
\label{sec:loss}
To facilitate end-to-end training of the entire model, we design a series of loss functions to constrain the learning process of parameters. In contrast to the original GCN approach~\cite{Li2022InteractingAG}, we augment our model with a localization module for both hands. This module incorporates an initialization scheme for the root node position, leveraging the hand center, and facilitates feature extraction at each central position. All our loss functions are provided in comprehensive detail below.

\noindent \textbf{Center Loss} is used to supervise our hand center learning. In essence, it is a pixel-wise binary logistic regression problem. The center points of the left and right hands are positive samples, while the rest are negative samples. Similar to CenterNet~\cite{zhou2019objects}, we use the form of focal loss~\cite{lin2020focal} to avoid the impact of imbalanced positive and negative sample sizes as follows:
\begin{equation}\label{eq3}
\mathcal{L}_{c} = \sum_{h \in \{L,R\}} (1 - A_h)^{\gamma} \log(A_h),
\end{equation}
where $A_h \in [0,1]$ is the estimated confidence map for the positive class, and $1- A_h$ is the probability for the negative class. $\gamma$ is a hyperparameter and is set to 2 in our experiment.

\noindent \textbf{Mask Loss} is used to supervise the generation of hand masks, which is a typical semantic segmentation problem. We use smooth $L_1$ loss to calculate the difference between prediction and ground truth.
\begin{equation}\label{eq3}
\mathcal{L}_{m} = ||M - \hat{M}||_{1},
\end{equation}
where $\hat{M}$ is the ground truth mask and $M$ is our mask prediction.

\noindent \textbf{Root Loss} represents the $L_1$ distance between the predicted root node and the ground truth. In this work, we select the first joint of the middle finger as our root node, which is the 9-th of the 21 joints.
\begin{equation}\label{eq3}
\mathcal{L}_{root} = \sum_{h \in \{L,R\}} ||Root^{h} - \hat{Root^{h}}||_{1}.
\end{equation}

\noindent \textbf{Mesh Loss} includes our GCN mesh loss and MANO mesh loss of 3D hand vertices, and we use $L_1$ loss for calculation.
{\small
\begin{equation}\label{eq3}
\mathcal{L}_{V} = \sum_{h \in \{L,R\}} ||\mathcal{M}^{h}_{GCN} -\hat{\mathcal{M}}^{h}_{GCN}||_{1} + 
||\mathcal{M}^{h}_{MANO} -\hat{\mathcal{M}}^{h}_{MANO}||_{1}.
\end{equation}
}
\noindent \textbf{Joint Loss.} We use the predefined joint regressor $\mathcal{J}$ in MANO to generate 3D joints. Similar to mesh loss, we use $L1_1$ loss.
\begin{equation}\label{eq3}
\mathcal{L}_{J} = \sum_{h \in \{L,R\}} || \mathcal{J} (\mathcal{M}^{h}_{MANO}) - \mathcal{J} (\hat{\mathcal{M}}^{h}_{MANO})||_{1}.
\end{equation}

\noindent \textbf{Re-projection Loss.} We use projection functions to project 3D meshes and key points onto 2D images to calculate the re-projection loss, which is achieved through $L_2$ loss.
\begin{equation}\label{eq3}
\begin{aligned}
\mathcal{L}_{rep} &= \sum_{h \in \{L,R\}} || (\Pi ( \mathcal{M}^{h}_{MANO}) - \Pi (\hat{\mathcal{M}}^{h}_{MANO}))||_{2} \\  
& + || (\Pi ( \mathcal{J}(\mathcal{M}^{h}_{MANO})) - \Pi (\mathcal{J} (\hat{\mathcal{M}}^{h}_{MANO})))||_{2}.
\end{aligned}
\end{equation}

\noindent \textbf{Smooth Loss.} To ensure the smoothness of the output mesh, we add normal vectors and edge length loss.
\begin{equation}\label{eq3}
\mathcal{L}_{smooth} = \sum_{i=1}^{3}|| e_{i} \cdot \hat{n}||_{1} + ||e - \hat{e}||_{1},
\end{equation}
where $\hat{n}$ and $e_{i}$ represent the ground truth normal vector and three edges on each face in the predicted mesh, respectively. $e$ represents the length of each edge, while $\hat{e}$ represents the corresponding ground truth.

\begin{table*}[t]
    \centering
    \caption{Comparison with previous SOTA methods on H2O~\cite{Hampali2021KeypointTS} evaluation dataset. We report the MPJPE/MPVPE and AL-MPJPE/AL-MPVPE (mm) for each hand.
    }
    \scriptsize 
    \begin{tabular}{c|cc|cccccccccccc}
        \toprule
        \multicolumn{1}{c}{\multirow{3}{*}{Methods}} &
        \multicolumn{2}{c}{Inputs} &
        \multicolumn{3}{c}{MPJPE$\downarrow$} & \multicolumn{3}{c}{MPVPE$\downarrow$} & 
        \multicolumn{3}{c}{AL-MPJPE$\downarrow$} &\multicolumn{3}{c}{AL-MPVPE$\downarrow$} \\

        \cmidrule(r){4-6} \cmidrule(r){7-9} \cmidrule(lr){10-12} \cmidrule(l){13-15} 
        \multicolumn{2}{c}{} RGB & D & Left h. & Right h. &  Mean h. & Left h. & Right h. &  Mean h. & Left h. & Right h. &  Mean h. & Left h. & Right h. &  Mean h. \\ \midrule
        H2O~\cite{Kwon2021H2OTH}  & \checkmark & & 41.45 & 37.21 & 39.33 & -  & - & - & - & - & - & - & - & -\\ 
        Hasson~\cite{hasson2020leveraging} & \checkmark & & 39.56 & 41.87 & 40.72 & -  & - & - & - & - & -& - & - & -\\ 
        Tekin~\cite{Tekin2019HOUE} & \checkmark & & 41.42 & 38.86 & 40.14 & -  & - & - & - & - & -& - & - & -\\ 
        HOI4D~\cite{Liu2022HOI4DA4} & \checkmark & & - & - & - & -  & - & - & 19.9 & 19.9 & 19.9 & - & - & -\\ 
        IntagHand~\cite{Li2022InteractingAG} & \checkmark & & 39.54 & 42.90 & 41.22 & 38.77  & 41.91 & 40.34 & 12.03 & 14.23 & 13.13 & 12.46 & 14.49 & 13.48\\
        kypt-trans~\cite{Hampali2021KeypointTS} & \checkmark & & -  & - & - & -  & - & - & 21.36 & 19.57 & 20.47 & - & - & -\\ 
        Ours-RGB & \checkmark & & 33.20 &36.28 & 34.74 & 33.00  & 35.55 & 34.28 & 10.95 & 12.71 & 11.83 & 11.28 & 12.90 & 12.09 \\
        \midrule
        
        PointNet++~\cite{Qi2017PointNetDH} &  & \checkmark & 17.17  & 17.83 & 17.50 & 16.98  & 17.37 & 17.18 & 7.61 & 9.40 & 8.51 & 7.82& 9.42 & 8.62 \\ 
        \midrule
        IntagHand+D~\cite{Li2022InteractingAG} & \checkmark & \checkmark & 17.26 & 20.92 & 19.09 & 16.79 & 20.20 & 18.49 & 9.82 & 12.56 & 11.19 & 10.10 & 12.56 & 11.33 \\
        kypt-trans+D~\cite{Hampali2021KeypointTS} & \checkmark & \checkmark & -  & - & - & -  & - & - & 15.58 & 14.50 & 15.04 & - & - & -\\ 
        Densefusion~\cite{Wang2019DenseFusion6O} & \checkmark  & \checkmark & 21.39  & 25.76 & 23.58 & 21.42 & 25.34 & 23.38 &18.08 & 23.55 & 20.81 & 18.42 & 23.79 & 21.11\\ 
        Ours-full  & \checkmark & \checkmark & \textbf{9.64} & \textbf{11.62} & \textbf{10.63} & \textbf{9.08} & \textbf{11.00} & \textbf{10.04} &  \textbf{6.93} & \textbf{8.74}  & \textbf{7.84} & \textbf{7.10} & \textbf{8.79} & \textbf{7.94} \\
        
        \bottomrule
    \end{tabular} 
    \label{tab:h2o}%
\end{table*}



\section{Experiment}
\subsection{Implementation Details}
Our proposed framework is implemented with PyTorch~\cite{Paszke2019PyTorchAI}, incorporating an asymmetric dual-stream architecture for feature encoding. Unlike IntagHand~\cite{Li2022InteractingAG} necessitating centered interactive hands in their training process, our model accommodates hands from arbitrary positions from the first perspective. For instance, with the H2O dataset~\cite{Kwon2021H2OTH} as our exemplar, we transform the input image into a square shape using zero padding and subsequently rescale it uniformly into $384 \times 384$. Although larger resolutions can preserve more details, they place higher demands on training memory. To conduct the training, we utilize two RTX2080Ti GPUs and assign a batch size of 8 instances per card. The initial learning rate is set to $1\times10^{-4}$ and decreases by a factor of 10 at the 30th epoch. The entire training procedure spans 80 epochs and typically takes approximately three days to complete. Common data augmentation strategies, including scaling, rotation, translation, color jittering, and horizontal flipping, are used during training.

\subsection{Datasets and Evaluation Metrics}


\noindent \textbf{H2O}~\cite{Kwon2021H2OTH} is a realistic two-handed dataset that contains multi-view RGB-D images. We only used the first perspective data, including 55,742 images in the training set, 11,638 images in the validation set, and 23,391 images in the test set. The dataset provides high-resolution images, with RGB images and depth maps being $1280 \times 720$ and pixel aligned. In addition, it provides 62-dimensional MANO annotations for each hand, which can generate corresponding 3D meshes and keypoints. With the help of the camera intrinsic matrix, we obtained the corresponding 2D landmarks. The dataset captures different objects in different desktop backgrounds, resulting in complex and varied hand poses.

\begin{figure*}[ht]
\centering
\includegraphics[width=0.98\textwidth]{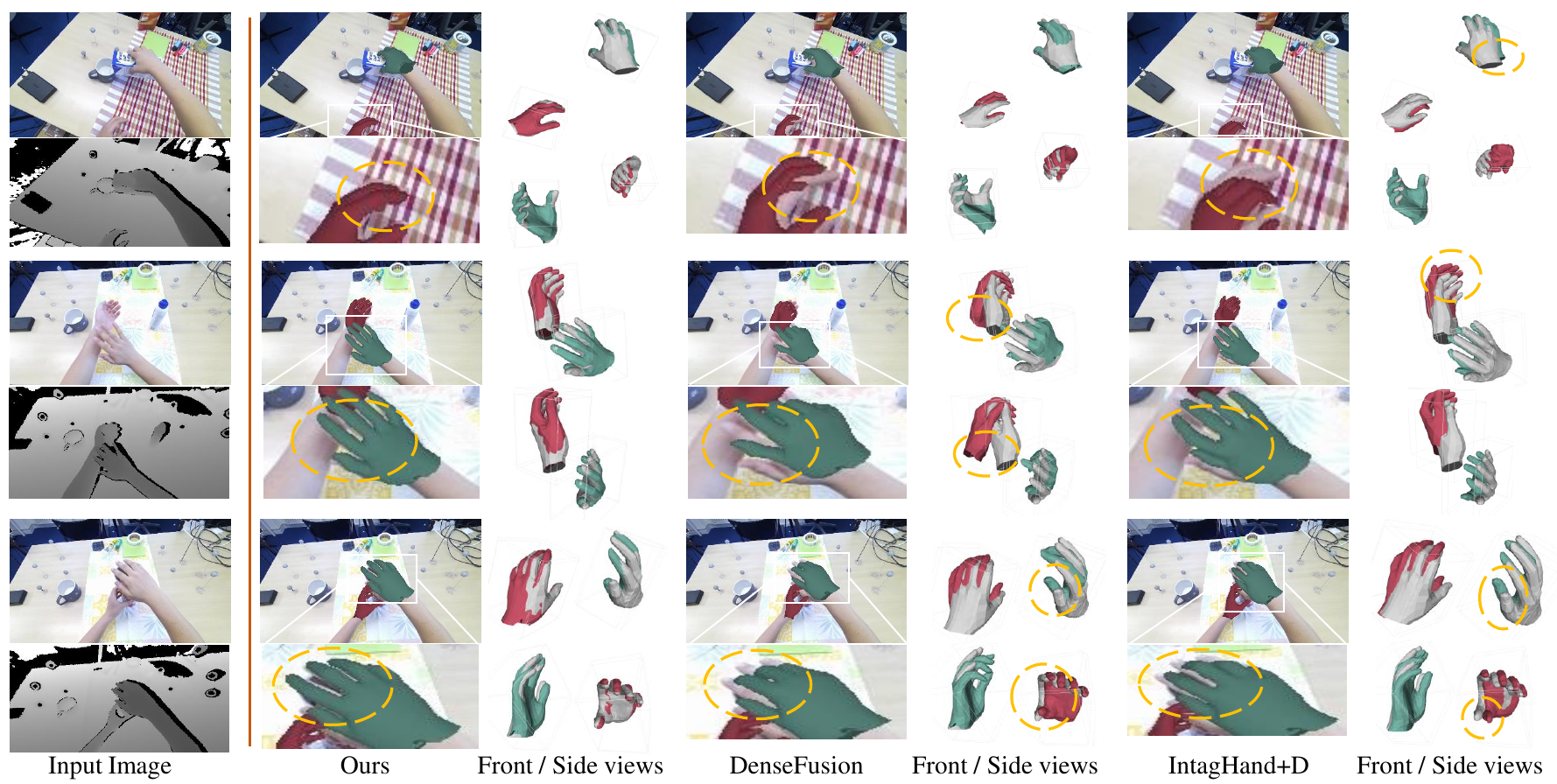} 
\caption{Visual comparison on the H2O~\cite{Hampali2021KeypointTS} dataset. We compared our results with DenseFusion~\cite{Wang2019DenseFusion6O} and IntagHand+D~\cite{Li2022InteractingAG}, and our results performed significantly better in hand-to-hand and hand-to-image alignment. We placed the predicted mesh and ground truth in the same coordinate system and color the left and right hands of the prediction in red and green respectively. From the side perspective, it can be seen that incorrect root node depth prediction can lead to significant misalignment.}
\label{visualH2O}
\vspace{-0.2in}
\end{figure*}

\noindent \textbf{H2O-3D}~\cite{Hampali2021KeypointTS} is a real captured dataset that focuses on the interaction scenarios between two hands and objects. There are a total of 17 multi-view sequences, with 5 experimenters participating and manipulating 10 different objects for recording. This dataset collected 3D annotations for 76,340 images, including 60,998 images from 69 single-camera sequences used in the training set and 15,342 images from 16 single-camera sequences used in the testing set. The dataset provides RGB-D image pairs with a resolution of $640 \times 480$, captured from a third perspective.

\noindent \textbf{RHD}~\cite{zimmermann2017learning} is a large-scale synthetic dataset redered from freely available characters. It provides 3D key points annotation for both hands from a third perspective, containing 41,258 training and 2,728 testing data. The RGB-D image pairs are pixel-aligned with resolution of $320 \times 320$.

\noindent \textbf{Evaluation Metrics.} To evaluate the accuracy of two-hand reconstruction, we used aligned mean per joint position error (AL-MPJPE) and aligned mean per vertex position error (AL-MPVPE) in millimeters to evaluate 3D key points and 3D mesh vertices after root node alignment, respectively. Practically, it becomes imperative to restore the accurate depth and scale of the reconstructed hands. Consequently, we additionally estimated the position of the root node and directly evaluated the MPJPE and MPVPE in the camera coordinate system. 

\subsection{Comparisons with State-of-the-art Methods}
\noindent \textbf{Two-hand reconstruction results on H2O dataset.} 
Firstly, we compared our method with previous SOTA two-hand reconstruction methods~\cite{Kwon2021H2OTH,Li2022InteractingAG,Liu2022HOI4DA4,Hampali2021KeypointTS} on H2O dataset. We also reported several single-hand pose estimation methods~\cite{hasson2020leveraging,Tekin2019HOUE}, using two separate models for the left and right hand images of H2O and the results reported in the table are borrowed from H2O~\cite{Kwon2021H2OTH}. Due to being the first method to use RGB-D input for two-hand reconstruction, there is few existing reconstruction method to compare. Therefore, we added depth input to existing RGB-based methods to demonstrate the superiority of our fusion strategy. Since DenseFusion~\cite{Wang2019DenseFusion6O} was originally designed for the 6-DoF pose estimation of objects, we integrated it into our proposed framework for comparison. Similarly, the original PointNet++~\cite{Qi2017PointNetDH} was designed for point cloud classification and segmentation, and we made corresponding modifications based on the authors' original implementation.
Table~\ref{tab:h2o} shows the evaluation results on H2O. Our method significantly surpasses the previous SOTA methods both in terms of absolute position error MPJPE/MPVPE and relative position error AL-MPJPE/AL-MPVPE. It obtains 9.64mm MPJPE of left hands and 11.62mm MPJPE of right hands under camera space. As for root-aligned position error, it obtains 6.93mm AL-MPJPE of left hands and 8.74mm AL-MPJPE of right hands. For a fair comparison, all methods in the table use the ground truth mask provided by the dataset to segment the depth map. In our subsequent ablation experiments, we also compared the results of directly using the mask estimated by the model. 

The visual comparison results on the testing set of H2O can be seen in Fig.~\ref{visualH2O}. We compared our method with DenseFusion~\cite{Wang2019DenseFusion6O} and IntagHand+D~\cite{Li2022InteractingAG} on the H2O testing set. By projecting the predicted meshes onto the input image, we can visually compare the alignment results of hand-to-image. In addition, we placed the ground truth meshes and prediction results simultaneously in the camera coordinate system to compare the alignment results of hand-to-hand. We circled the parts with obvious differences in yellow in Fig.~\ref{visualH2O}. It can be seen that our method achieved significantly better alignment in all perspectives.

\begin{figure*}[ht]
\centering
\includegraphics[width=0.9\textwidth]{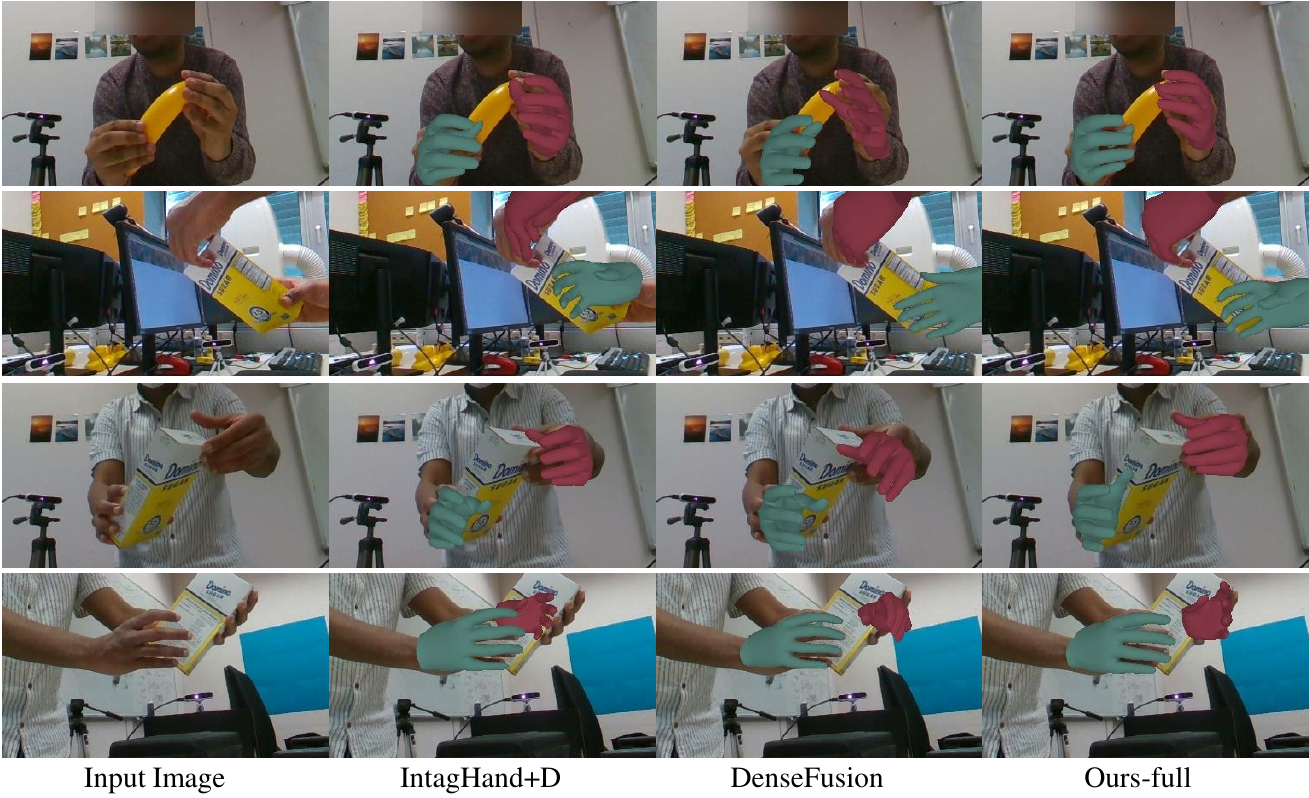} 
\caption{Visualization results on the H2O-3D~\cite{Hampali2021KeypointTS} test set. We compared our results with IntagHand+D~\cite{Li2022InteractingAG} and DenseFusion~\cite{Wang2019DenseFusion6O}, and our results performed significantly better in hand-to-image alignment.}
\label{h2o3d}

\end{figure*}

\begin{table}[t]
    \centering
    \caption{Comparison with previous SOTA methods on H2O-3D~\cite{Hampali2021KeypointTS} evaluation dataset. We report the AL-MPJPE (mm) and AUC for each hand.
    }
    \scriptsize 
    \begin{tabular}{c|cc|cc}
        \toprule
        \multicolumn{1}{c}{\multirow{2}{*}{Methods}} &
        \multicolumn{2}{c}{Inputs} &
        \multicolumn{1}{c}{\multirow{2}{*}{AL-MPJPE$\downarrow$}} & 
        \multicolumn{1}{c}{\multirow{2}{*}{AUC$\uparrow$}} 
        \\
        \multicolumn{2}{c}{} RGB & D \\
\midrule
        Densefusion~\cite{Wang2019DenseFusion6O} & \checkmark & \checkmark  &  12.7 & 0.747 \\  
        IntagHand+D~\cite{Li2022InteractingAG} & \checkmark & \checkmark  &  11.7 & 0.767 \\  
        kypt-trans+D~\cite{Hampali2021KeypointTS} & \checkmark & \checkmark  &  11.7 & 0.769 \\  
        Ours-full & \checkmark & \checkmark  &  \textbf{10.7} & \textbf{0.787} \\  
        \bottomrule
    \end{tabular} 
    \label{tab:h2o3d}%
\end{table}

\noindent \textbf{Two-hand reconstruction results on H2O-3D dataset.} H2O3D is a challenging two-hand dataset from a third perspective. Our model has also been tested separately on it. As the dataset did not provide annotations for the testing set, we submitted and evaluated it on the official online platform\footnote{https://codalab.lisn.upsaclay.fr/competitions/4897}. The evaluation results are shown in Table \ref{tab:h2o3d}, and it can be seen that our method has achieved better performance under the specified evaluation indicators of the dataset. The visualization results are shown in Fig.~\ref{h2o3d}.


\begin{table*}[!ht]
\caption{Ablation study using different experimental configurations evaluated on H2O test set. Checking FTN indicates the use of the dedicated feature space transformation network designed in this article while unchecking it indicates the use of simple feature concatenating. Checking GT Mask means using a ground truth mask to segment the depth map while unchecking it means using the mask predicted by the model.}
\centering
\resizebox{0.75\textwidth}{!}{
\begin{threeparttable}
\begin{tabular}{c|cc|ccc|cc|cccc}
\toprule
    & RGB & Depth & Point Feat. & RGB Feat. & Center Feat. & FTN & GT Mask & MPJPE$\downarrow$ & MPVPE$\downarrow$ & AL-MPJPE$\downarrow$ & AL-MPVPE$\downarrow$ \\
\midrule
    1& $\checkmark$ & - & - & - & $\checkmark$  & - & -&34.74 & 34.28 & 11.83  & 12.09 \\
    2& - & $\checkmark$ & $\checkmark$ & - & -  & - & $\checkmark$& 17.50 & 17.18 & 8.50 & 8.62  \\
    3& $\checkmark$ & $\checkmark$ & - & - & $\checkmark$  & -& - & 16.15 & 15.70 & 10.73 & 10.88  \\
\midrule 
    4& $\checkmark$ & $\checkmark$ &$\checkmark$ & $\checkmark$ & $\checkmark$ & - & $\checkmark$ &  16.40 & 15.99 & 8.32 & 8.41  \\   
    5& $\checkmark$ & $\checkmark$ & $\checkmark$& $\checkmark$ & -  & $\checkmark$ & $\checkmark$ & 12.61 & 12.32 & 8.53 & 8.62  \\
    6& $\checkmark$ & $\checkmark$ & $\checkmark$& $\checkmark$ & $\checkmark$  & $\checkmark$ & - & 12.11 & 11.62 & 8.89 & 8.96  \\
    7& $\checkmark$ & $\checkmark$ & $\checkmark$ & $\checkmark$ & $\checkmark$  & $\checkmark$ & $\checkmark$ & \textbf{10.63} & \textbf{10.04} & \textbf{7.84}  & \textbf{7.94} \\
\bottomrule
\end{tabular}
\end{threeparttable}}
\vspace{-0.12in}
\label{tab:ablation}%
\end{table*}

\begin{table}[t]
    \centering
    \caption{Ablation studies using different decoders evaluated on H2O~\cite{Hampali2021KeypointTS} test set.
    }
    \scriptsize 
    \begin{tabular}{ccccc}
        \toprule
        Decoders & MPJPE$\downarrow$ & MPVPE$\downarrow$ & AL-MPJPE$\downarrow$ & AL-MPVPE$\downarrow$ \\
\midrule
        MANO & 16.60  & 16.69 & 8.36 & 8.54 \\  
        GCN & \textbf{10.63} & \textbf{10.04} & \textbf{7.84}  & \textbf{7.94} \\
        
        \bottomrule
    \end{tabular} 
    \label{tab:decoders}%
\end{table}

        

        

\begin{table}[t]
    \centering
    \caption{Ablation study using different number of pyramid layers evaluated on H2O dataset.
    }
    \scriptsize 
    \begin{tabular}{ccccc}
        \toprule
        Num. of Layers & MPJPE$\downarrow$ & MPVPE$\downarrow$ & AL-MPJPE$\downarrow$ & AL-MPVPE$\downarrow$ \\
 
\midrule
        L1  & 21.97  & 21.56 & 8.44 & 8.63  \\  
        L2  & 19.58  & 19.11 & 8.95 & 9.12  \\  
        L3  & 11.76  & 11.28 &8.03 &8.15  \\  
        L1+L3  & 11.47  & 11.05 &8.43 &8.58  \\  
        L1+L2+L3  & \textbf{10.63} & \textbf{10.04} & \textbf{7.84}  & \textbf{7.94} \\
        
        \bottomrule
    \end{tabular} 
    \label{tab:pyramidlayers}%
\end{table}

\subsection{Ablation Study}
We conducted a series of extensive ablation experiments on H2O and RHD datasets to confirm the contributions of different modules within our framework. Firstly, our objective is to demonstrate the performance improvement achieved by incorporating depth maps comparing to using solely RGB inputs. This aligns with the intuition of most individuals and our original intention behind the design of fusion modules. Secondly, we aim to demonstrate the advantage of our proposed PDFNet algorithm over existing fusion strategies. To achieve this, we focus on modifying only the feature fusion module within the same encoder-decoder framework. Therefore, we show the efficacy of our pyramid design and feature transformation module. In the following, we present thorough ablation experiments and provide a detailed analysis of the corresponding outcomes.

\begin{figure}[ht]
\centering
\includegraphics[width=0.98\columnwidth]{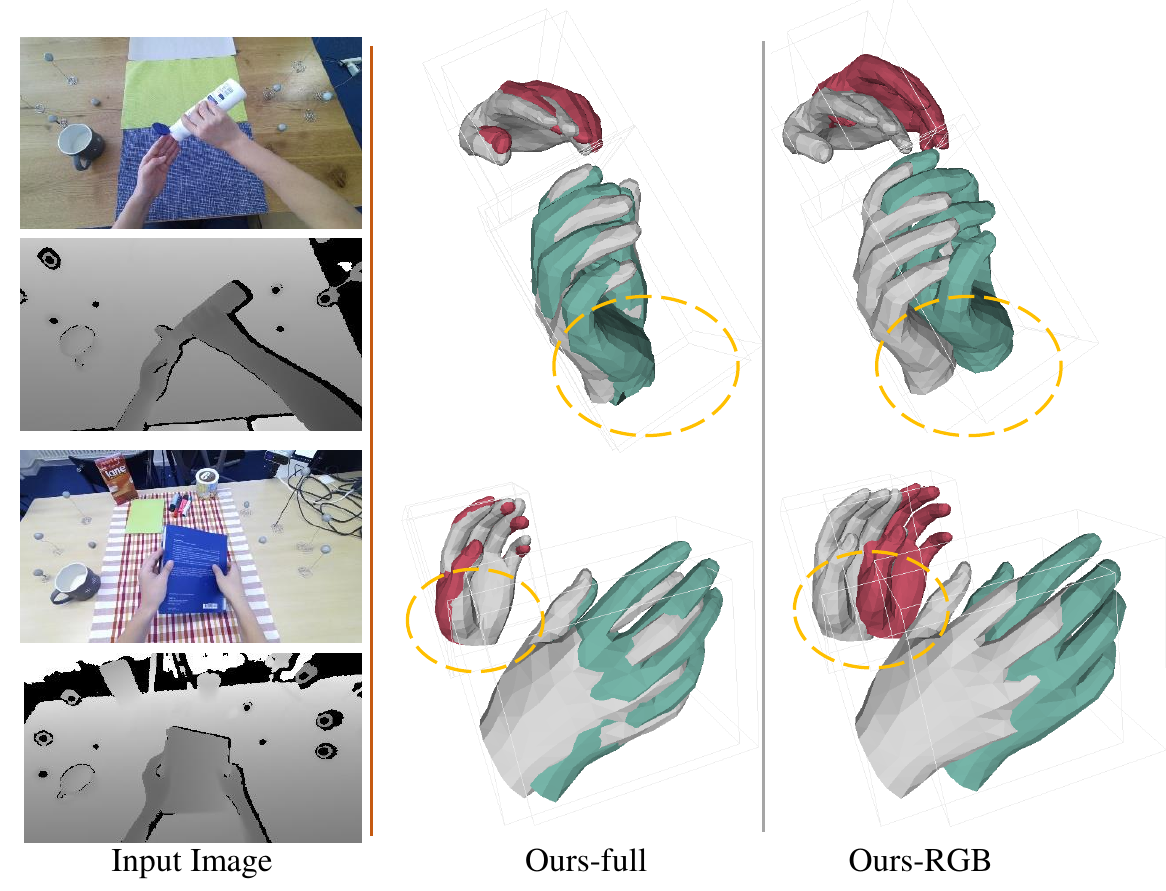} 
\caption{Visual comparison of Ours-full model with Ours-RGB model in Table~\ref{tab:h2o} on the H2O~\cite{Hampali2021KeypointTS} dataset. We placed the predicted meshes and ground truth in the same camera space and color the predicted left and right hands with red and green respectively.}
\label{depthcompare}
\end{figure}

\noindent \textbf{Comparison of different input modalities.}
We conducted a comprehensive analysis of our model's performance under different input scenarios, and the experimental results evaluated on H2O dataset are presented in Table~\ref{tab:ablation}. Row-1 in the table denotes the baseline method, where only RGB images are employed for feature extraction. In Row-2, only depth maps are utilized to extract point features, with the aid of ground truth masks to generate the initial point cloud. Row-3 illustrates the model's direct usage of 4-channel RGB-D images as input. Comparing rows 1-3 in the table, it becomes evident that the integration of depth maps leads to a significant reduction in error by $50\%$. This aligns with our initial expectations, as predicting depth solely from RGB input inherently presents challenges due to the ill-posed nature of the problem. Visual comparisons are showcased in Fig.~\ref{depthcompare}.

Row-3 attains lower absolute position error compared to Row-2, however, it does not possess an advantage in terms of relative position error. This indicates that RGB images contribute to improving the accuracy of the final predictions while also introducing some background interference information. Moreover, it signifies that a simplistic and coarse 4-channel input is not an ideal solution. This emphasizes our preference to fully leverage the complementary nature of the two input modalities.

\noindent \textbf{Comparison of different fusion strategies.}
By comparing the results of DenseFusion~\cite{Wang2019DenseFusion6O} and Ours-full in Table~\ref{tab:h2o}, it can be found that our devised pyramid feature fusion method confers significant performance advantages. The absolute position error (MPJPE) has decreased from 23.58mm to 10.63mm, while the relative position error (AL-MPJPE) has decreased from 20.81mm to 7.84mm. 

Furthermore, we delved further into the impact of different modules in PDFNet on the final performance, as shown in Table~\ref{tab:ablation}. Through a comparison between Row-4 and Row-7, we observe that the feature transformation network (FTN) plays a pivotal role in reducing model error, underscoring the necessity of adaptive weight allocation for the two input modalities. Otherwise, the introduction of undesired background interference features, as seen in Row-3, would adversely affect the final model performance.

In Row-5, we removed the center feature and directly utilized the fusion result of the point feature and RGB feature as the final feature. It is noteworthy that this approach leads to an increase of approximately 2mm in MPJPE and 1mm in AL-MPJPE compared to the full model. We posit that incorporating global center features proves advantageous, as it prevents the model from falling into local optima or overfitting by effectively leveraging the informative global-local features.

In Row-6, we attempted to employ the model's estimated mask for segmenting the depth map. However, this resulted in an increase of nearly 1mm in all error metrics. This phenomenon can be attributed to the inherent challenges in semantic segmentation itself, where there is inevitably a disparity between the predicted mask and the ground truth. Luckily, our model's performance only experienced a minor decrease while still surpassing the state-of-the-art (SOTA) methods significantly.

In addition, we also explored the impact of different pyramid layers on feature fusion and model performance on the H2O dataset. The experimental results are shown in Table~\ref{tab:pyramidlayers}. It can be seen that by integrating pyramid features, stronger representation ability can be achieved, which also improves the final performance of the model.

\noindent \textbf{Comparison of different decoders.}
To demonstrate the efficacy of our framework and the ease deployment of PDFNet, we replaced our GCN-based decoder with a MANO-based decoder. The experimental results are shown in Table~\ref{tab:decoders}, indicating that the GCN module used in our framework has achieved significant performance advantages. In addition, compared to the results of IntagHand and IntagHand+D in Table~\ref{tab:h2o}, our MANO version still achieved improvements of 24.62mm MPJPE and 2.49mm MPJPE, respectively. This indicates that our PDFNet is an effective feature fusion algorithm that can extract more effective features for MANO-based decoders.

Besides, we use the RHD dataset to test whether our PDFNet module can work in complex scenarios, as this dataset is a synthetic two-hand dataset from a third perspective. It is very challenging because the position of the hand varies greatly, from a very close range of over 10mm to a distance of over 2 meters. Our results in Table~\ref{tab:rhd} confirm the robustness of our PDFNet. Besides, as this dataset only provides annotations for sparse 3D key points, we replaced the GCN-based decoder with a simple three-layer fully connected layer to directly output the coordinates of 21 key points. This experiment also proves that our PDFNet algorithm has a certain universality and benefits multiple decoders.

\begin{table}[t]
    \centering
    \caption{Performance comparison between using the PDFNet module and not using it on the RHD~\cite{zimmermann2017learning} test set. We report the MPJPE and AL-MPJPE (mm) for each hand here.
    }
    \scriptsize 
    \begin{tabular}{c|cccc}
        \toprule
        \multicolumn{1}{c}{\multirow{2}{*}{Methods}} &
        \multicolumn{2}{c}{MPJPE$\downarrow$} & \multicolumn{2}{c}{AL-MPJPE$\downarrow$} \\
        \multicolumn{1}{c}{} & Left h. & Right h. & Left h. & Right h. \\ 
        \midrule
        w/o PDFNet & 419.89  & 451.03 & 55.81 & 50.81 \\  
        w/ PDFNet & \textbf{215.34} & \textbf{218.29} & \textbf{36.90}  & \textbf{35.99} \\
        
        \bottomrule
    \end{tabular} 
    \label{tab:rhd}%
\end{table}


\noindent \textbf{Limitations.}
The precision and generalizability of the model's mask predictions are not yet optimal. It is worthy of contemplating the utilization of expansive pre-trained models, such as SAM~\cite{kirillov2023segany}, to achieve greater adaptability across a wider application scenarios. In real-world implementations, the incorporation of temporal information from consecutive frames is imperative in acquiring consistent estimations. Regrettably, our current methodology solely supports single-frame RGB-D images as input, indicating room for further improvement.

\section{Conclusion}
This paper presents a comprehensive end-to-end framework for reconstructing both hands from a single RGB-D input. We adopt a well-designed dual-stream architecture to extract depth and RGB features, separately. Moreover, a novel pyramid feature fusion algorithm, named PDFNet, is introduced to synergistically leverage the strengths of these two complementary input modalities. The model successfully generates dense two-hand meshes in the camera coordinate system by employing our GCN-based decoder. Experiments have shown that the fusion algorithm and reconstruction framework proposed in this paper can accurately reconstruct two-hand meshes with real depth and scale. Compared to the state-of-the-art methods, our approach obtains a remarkable enhancement in performance. In future work, we aim to explore hand-object interaction and human-environment interaction to broaden the scope of application scenarios. Furthermore, both temporal and multi-perspective information can be considered to improve the usability of the model.

\bibliographystyle{IEEEtran}
\bibliography{mybib}


\begin{IEEEbiography}[{\includegraphics[width=1in,height=1.25in,clip,keepaspectratio]{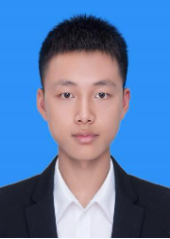}}]{Jinwei Ren} is currently a PhD candidate in the College of Computer Science and Technology,
Zhejiang University, Hangzhou, China. Before that, he received the bachelor degree from Chongqing University, China, in 2017. His research interests
include SLAM and computer vision, with
a focus on 3D reconstruction.
\end{IEEEbiography}

\begin{IEEEbiography}[{\includegraphics[width=1in,height=1.25in,clip,keepaspectratio]{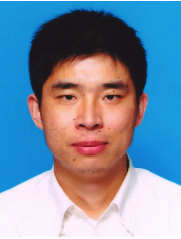}}]{Jianke Zhu}{\,} received the master’s degree from University of Macau in Electrical and Electronics Engineering, and the PhD degree in computer science and engineering from The Chinese University of Hong Kong, Hong Kong in 2008. He held a post-doctoral position at the BIWI Computer Vision Laboratory, ETH Zurich, Switzerland. He is currently a Professor with the College of Computer Science, Zhejiang University, Hangzhou, China. His research interests include computer vision and multimedia information retrieval. 
\end{IEEEbiography}

\vspace{11pt}


\vfill

\end{document}